\documentclass[runningheads]{llncs}
\usepackage{graphicx}
\graphicspath{{pdf_figures/}}
\DeclareGraphicsExtensions{.pdf}
\usepackage{amsmath}
\usepackage{subfig}
\usepackage{url}
\usepackage{epsfig}
\usepackage{amssymb}
\usepackage[utf8]{inputenc}
\usepackage[english]{babel}
\usepackage [autostyle, english = american]{csquotes}
\MakeOuterQuote{"}

\usepackage{algorithm,tabularx}
\usepackage{algpseudocode}

\usepackage{multirow}
\usepackage{makecell}

\usepackage{xspace}

\usepackage{color}
\usepackage{tikz}

\algrenewcommand\algorithmicrequire{\textbf{Input:}}
\algrenewcommand\algorithmicensure{\textbf{Output:}}

\makeatletter
\newcommand{\multiline}[1]{%
	\begin{tabularx}{\dimexpr\linewidth-\ALG@thistlm}[t]{@{}X@{}}
		#1
	\end{tabularx}
}
\makeatother

\newcounter{algsubstate}

\makeatletter
\algnewcommand{\ProperState}[1]{\Statex \hskip\ALG@thistlm #1}
\makeatother

\makeatletter
\def\input@path{{pdf_figures/}}
\makeatother

\DeclareMathOperator{\EX}{\mathbb{E}} 
\DeclareMathOperator{\loss}{\mathcal{L}} 

\newcommand{\SSIM}{\mathrm{SSIM}}
\newcommand{\GAC}{\mathrm{GAC}}
\newcommand{\GARDiN}{\mathrm{GARDiN}}

\newcommand{\FL}{\mathrm{FL}}

\makeatletter
\DeclareRobustCommand\onedot{\futurelet\@let@token\@onedot}
\def\@onedot{\ifx\@let@token.\else.\null\fi\xspace}

\makeatother

\newcolumntype{P}[1]{>{\centering\arraybackslash}p{#1}}

\begin{document}
\title{Local Anomaly Detection in Videos using Object-Centric Adversarial Learning\thanks{Supported by grants from IVADO and NSERC funding programs.}}
\titlerunning{Local Anomaly Detection in Videos}
%
\author{Pankaj Raj Roy\inst{1} \and
Guillaume-Alexandre Bilodeau\inst{1} \and
Lama Seoud\inst{2}}
\authorrunning{Roy et al.}
%
\institute{LITIV, Dept. Computer and Software Engineering, Polytechnique Montr\'{e}al\\
\email{\{pankaj-raj.roy,gabilodeau\}@polymtl.ca} \and
Institute of Biomedical Engineering, Polytechnique Montr\'{e}al\\
\email{lama.seoud@polymtl.ca}}
\maketitle              
\begin{abstract}
We propose a novel unsupervised approach based on a two-stage object-centric adversarial framework that only needs object regions for detecting frame-level local anomalies in videos. The first stage consists in learning the correspondence between the current appearance and past gradient images of objects in scenes deemed normal, allowing us to either generate the past gradient from current appearance or the reverse. The second stage extracts the partial reconstruction errors between real and generated images (appearance and past gradient) with normal object behaviour, and trains a discriminator in an adversarial fashion. In inference mode, we employ the trained image generators with the adversarially learned binary classifier for outputting region-level anomaly detection scores. We tested our method on four public benchmarks, UMN, UCSD, Avenue and ShanghaiTech and our proposed object-centric adversarial approach yields competitive or even superior results compared to state-of-the-art methods.

\end{abstract}
%
%
%
\section{Introduction}
Detecting anomalies in surveillance videos allows designing safer living environments by identifying potential risks, unsafe interactions between users or confusing urban signage. Similarly to previous work~\cite{Lu2013,Ribeiro2018,Wang2018b}, we define abnormal event detection as the identification of spatio-temporal image regions in a video that deviate from the learned normal ones. We focus on detecting abnormal events on a per individual/object basis, also known as local anomalies. They are independent from other surrounding spatio-temporal events. Thus, we only need to consider the image regions corresponding to objects possessing the abnormal behaviour. We want to detect those events with just a small number of frames.

\begin{figure}[t]
	\centering
	\def\svgwidth{0.6\linewidth}
	\input{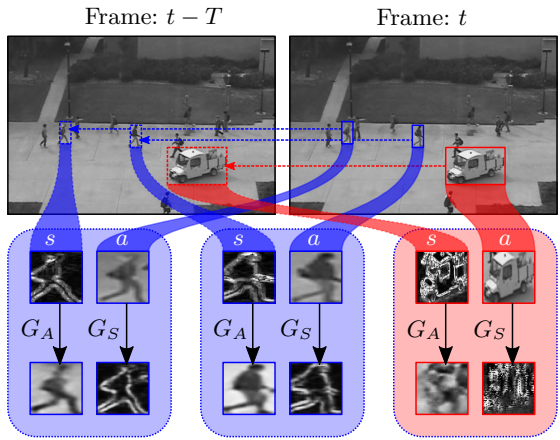}
	\caption{Generated images from our proposed method. $ G_{S} $ generates the past spatial gradient image $ s $ from the appearance $ a $ of a region in a frame $ t $ and $ G_{A} $ does the reverse. For an abnormal region (in red), the images are not generated correctly compared to the normal regions (in blue).}
	\label{fig:our_method}
\end{figure}

Recently, a solution based on a Convolutional Auto-Encoder (CAE) was proposed in~\cite{Ionescu2019} for detecting local anomalies, which takes less memory for building the networks and for which the training is significantly faster compared to holistic methods that consider the whole image, not just objects' bounding boxes. This is seen as an object-centric approach, due to the fact that it ignores background information and learns to classify local anomalies solely based on local information of the objects. However, it relies on K-means clustering combined with a one-versus-rest SVM classification scheme, that requires a predefined knowledge on the number of clusters which might vary depending on the scenario. Moreover, the CAE models are trained separately and are, therefore, unable to learn the relation between different local information like appearance and gradient.

To tackle these issues, we propose a new method for detecting local anomalies in videos that uses a novel unsupervised two-staged object-centric adversarial framework. The first stage of our method learns the normal local gradient-appearance correspondences and the second stage learns to classify events in an unsupervised manner. The local gradient-appearance correspondence is learned by relating the gradients of a previous frame with the visual appearance of an object in the current frame.

Our method first uses a pretrained object detector to extract all the regions of interest in a frame, and then, extracts the spatial gradients in the previous frame at the location of the detected objects in the current frame. After that, we train the components of our generative framework: 1) two cross-domain generators, where one learns to predict the past gradients by taking the appearance and the other one learns the reverse, and 2) two discriminators that discriminate between the real and generated appearance and real and generated gradients, respectively. This first stage results in the training of two cross-domain transformers. For the second stage, we apply the cross-domain transformers for generating gradient and appearance images from the normal real appearance and gradient images. Then, we compute the partial mean-squared reconstruction errors (PMSRE) between real and the generated images and train a generative adversarial network (GAN) with a generator, which generates fake PMSRE, and a discriminator, a binary classifier, that determines whether they are real or not. 

During inference, when an anomaly occurs as illustrated in the figure~\ref{fig:our_method}, one or both of the transformers ($ G_{A} $ and $ G_{S} $) will not be able to correctly predict the past spatial gradients or/and the current appearance, thus, indicating the likelihood of that region of being abnormal. We tested our proposed approach on four public datasets with local anomalies: UMN~\cite{Mehran2009}, UCSD~\cite{Mahadevan2010}, Avenue~\cite{Lu2013} and ShanghaiTech~\cite{Luo2017}. The results show that our method is better or competitive with the state-of-the-art on all datasets.

Our contributions are the following: 1) We propose a novel two-stage object-centric adversarial approaches for local anomaly detection in videos, 2) we employ an unsupervised cross-domain GAN trained using pixel-level regions of the objects having normal behaviour in videos and 3) we propose an adversarially-learned binary classifier that classifies normal from abnormal PMSRE.

\section{Related Work}
Before the success of deep learning methods, most of the authors were relying on manually predefined feature extraction. For example, in \cite{Mousavi2015}, the authors extract histograms of oriented tracklets from a few consecutive frames. Features derived from optical flow are also used for detecting abnormal events through the use of a covariance matrix \cite{Wang2018b}. The success of these methods to detect abnormal events depends on the quality of the extracted hand-crafted features, and thus, the quality of the detection is heavily influenced by them. Besides, considering engineered features instead of pixel data to learn normal/abnomal classification implies loss of valuable spatial and temporal information.

Instead of analyzing image regions, another approach is to classify as normal or abnormal foreground object trajectories. For instance, in \cite{Li2013}, the authors compute trajectories for normal events, apply sparse reconstruction analysis on them to learn the normal patterns, and detect any abnormal trajectories as outliers. For abnormal trajectory detection of road users, an unsupervised approach via a deep Auto-Encoder (DAE) was proposed in \cite{Roy2018a} for learning the normal trajectories and detecting the abnormal ones as outliers. In a following work~\cite{Roy2019a}, the use of a GAN in a discriminative manner was applied for classifying abnormal trajectory reconstruction errors produced by a pretrained DAE. This inspired us for applying an adversarial approach for detecting abnormal events in videos. Despite the fact that the problem becomes simpler when converting events into trajectories, this approach suffers from the fundamental issue of losing appearance information and relying on an external mechanism for obtaining trajectory data. Normal trajectories are also scene specific. 


The authors of \cite{Ribeiro2018} proposed to use a CAE to learn the normal appearance and motion features extracted using a Canny edge detector and optical flow. Once the CAE is fully trained, it is then used on every frame for constructing the regularization of reconstruction errors (RRE), which will later be used for detecting anomalies. Nevertheless, this approach trained solely on the normal samples might start to generalize over the abnormal ones, thus affecting the classification performance. Recently, an object-centric approach using CAE models was proposed in \cite{Ionescu2019} in which the generated latent appearance and motion features (motion features are actually computed from past and future gradient images) of objects are used for classifying local anomalies. However, the CAE models, one for appearance and two for motion, are trained separately and several SVM classifiers are required for doing the anomaly classification. We took inspiration from this work by applying the object-centric input images to the framework of a cross-domain GAN proposed in \cite{Kim2017}, which allows us to better learn the gradient and appearance correspondence by jointly training two image generators in an end-to-end manner. We also improve classification using adversarial learning of reconstruction errors following \cite{Roy2019a}. Moreover, we only relies on the previous frame for gradient, thus making it applicable in real-time scenarios.

To solve the aforementioned issues with CAEs, some authors proposed to use GANs for training a discriminator in an unsupervised fashion by using the generator to generate abnormal data during the learning process \cite{Sabokrou2019}. The trained discriminator can then later be used as a binary classifier for detecting anomalies. However, these methods cannot handle the spatio-temporal aspect in the video, and thus, perform poorly when the appearance change over time. To tackle this problem, authors in \cite{Liu2018,Nguyen2019a} proposed to use a GAN that can learn to produce the future frame of the scene with normal events, and, when an abnormal event occurs in a scene, the generator will not produce the correct subsequent frame, thus allowing the detection of abnormal events. The downside of these methods is that they rely on the optical flow methods and cannot be generalized across different scenes.

\section{Proposed Method}
In order to alleviate the shortcomings of existing methods, our proposed approach incorporates an object-centric mechanism into an adversarially-learned prediction-based method, to learn to distinguish local anomalies based on the appearance/gradient of regions of object of interest, thus ignoring background information.

Our method first detects all the objects in a video frame by using a multiclass object detector, and extracts the spatial gradient image of the corresponding regions of interest. Then, we train a GAN, called GARDiN (Gradient-Appearance Relation Discovery Network) inspired from DiscoGAN~\cite{Kim2017} to discover the object-centric cross-domain relations between past spatial gradients (that capture shapes and patterns) and the current visual appearance using the extracted data. This allows GARDiN to learn how the shape/appearance of a region evolves over time. This jointly trains two cross-domain object-centric generators that each transforms an image from one domain (appearance/gradient) into another domain (gradient/appearance), and two discriminators that each learns to discriminate a specific real domain against the generated fake one. After that, we construct normal partial mean squared reconstruction errors (PMSRE) produced by comparing the real appearance and gradient images of the regions and the generated ones. Inspired by \cite{Roy2019a}, we then use these PMSRE for training a discriminator that acts as a binary reconstruction error classifier through a typical GAN-based approach in which the generator learns to generate realistic PMSRE, while the discriminator discriminates them from the real ones. 

Once fully trained, we apply GARDiN to obtain PMSRE and directly use our adversarially-learned PMSRE classifier to discover whether the object-centric region is normal or not. We named our anomaly detection system GARDiN video anomaly detector (GARDiN-VAD).

\subsection{Object Detection and Gradient Extraction}
In order to detect multiple objects in a video frame, we use the pretrained multiclass object detector, CenterNet~\cite{zhou2019objects}, which is currently one of the best and readily available in machine learning frameworks. This detector is both reasonably accurate and fast enough for an anomaly detection system. Also, since it does not rely on implicit anchors, it can detect well small objects, which is crucial for detecting anomalies in a crowded scene. Note that, to obtain the appearance images, we transform all the detected objects into grayscale and resize them into $ 64 \times 64 \times 1 $.

In addition to detecting the spatial locations of objects frame by frame in a given video, we also compute an object-centric past spatial gradient image for each object as in \cite{Ionescu2019}, which is defined as the 2D spatial gradient produced by the Sobel operator on the region in the previous frame using the bounding box coordinates of the object in the current frame. This past spatial gradient image enables our adversarial framework to implicitly learn the change of object shape and position as the object moves. Moreover, compared to an optical flow image, it is significantly less expensive to compute and generalizes well, as it ignores the specific motion direction, thus facilitating the unsupervised learning of normal motion patterns. Note that, because the change caused by motion is small in two consecutive frames in a 25 fps rate video, we use a temporal spacing of $T$ frames when computing the past spatial gradient image ($T=3$ in our experiments).

\subsection{Gradient-Appearance Relation Discovery Network (GARDiN)}
Inspired by cross-domain GAN~\cite{Kim2017} that discover the relationship between images across different domains, we propose to apply this idea for learning the correspondence between the appearance and the gradient of an object moving in a video. Thus, we define appearance and gradient as two distinct domains in which the goal is to discover the relationship between images belonging to each of these domains.

\begin{figure}[t]
	\centering
	\def\svgwidth{0.7\linewidth}
	\input{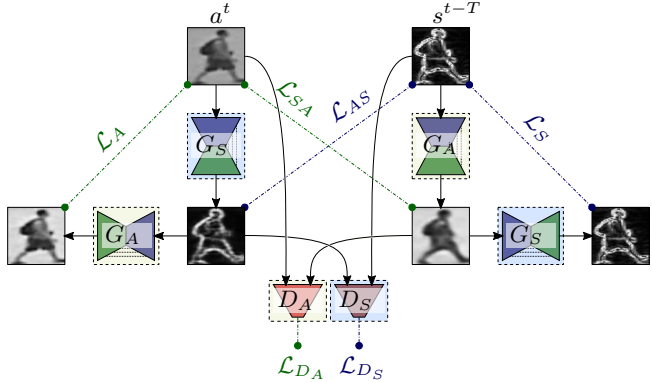}
	\caption{Adversarial framework of GARDiN. During the training process, for a frame $ t $, $ G_{S} $ and $ G_{A} $ learn to generate images $ s^{t-T} $ and $ a^{t} $ respectively from spatial gradient domain $ S $ and  from appearance domain $ A $ by using the reconstruction losses $ \loss_{A} $, $ \loss_{S} $, $ \loss_{AS} $ and $ \loss_{SA} $, while the appearance discriminator $ D_{A} $ and spatial gradient discriminator $ D_{S} $ classify the real against the generated ones with $ \loss_{D_{A}} $ and $ \loss_{D_{S}} $.}
	\label{fig:our_method_discoman}
\end{figure}

\subsubsection{Formulation}
As illustrated in figure \ref{fig:our_method_discoman}, our GAN is composed of two generators that either transform an appearance image into a gradient image or a gradient image into an appearance image, and two discriminators that each discriminates the real appearance/gradient image against the transformed one. More specifically, considering the two domains, appearance $ A $ and spatial gradient $ S $, the generator $ G_{S} $ maps the images from $ A $ to $ S $ and generator $ G_{A} $ from $ S $ to $ A $. For instance, in a video frame $ t $, given an input appearance image $ a^{t} $ from domain $ A $ of an object, $ G_{S} \left( a^{t} \right) $ should produce an image resembling the real corresponding spatial gradient image $ s^{t-T} $ from domain $ S $. In addition to that, $ G_{A} \left( G_{S} \left( a^{t} \right) \right) $ should reproduce the original input image $ a^{t} $. The same applies for an input spatial gradient image $ s^{t-T} $ from domain $ S $. Therefore, for a given video frame $ t $, we can formulate four reconstruction losses:
\begin{align} \label{equ:reconst_losses}
	\begin{split}
		\loss_{AS} &= d \left( G_{S} \left( a^{t} \right), s^{t-T} \right) \\
		\loss_{SA} &= d \left( G_{A} \left( s^{t-T} \right), a^{t} \right) \\
		\loss_{A} &= d \left( G_{A} \left( G_{S} \left( a^{t} \right) \right), a^{t} \right) \\
		\loss_{S} &= d \left( G_{S} \left( G_{A} \left( s^{t-T} \right) \right), s^{t-T} \right) \textnormal{,}
	\end{split}
\end{align}
where $d()$ is a custom distance function, $ \loss_{AS} $ and $ \loss_{SA} $ deal with the transformation of an image from $ A \rightarrow S $ and from $ S \rightarrow A $ respectively, $ \loss_{A} $ and $ \loss_{S} $ apply to the reconstruction of the given input image using the two generators in sequence. Empirically, we found that combining different distance measures such as L1, L2 and SSIM~\cite{Wang2004} yields the best performance, as demonstrated in our ablation study. Thus, the distance $ d(I_1,I_2) $ between two images $ I_1 $ and $ I_2 $ is given by the following equation:
\begin{equation}
	d \left( I_1, I_2 \right) = d_{L1} \left( I_1, I_2 \right) + d_{L2} \left( I_1, I_2 \right) + d_{ss} \left( I_1, I_2 \right)
\end{equation}
where
\begin{align}
	\begin{split}
		d_{L1} \left( I_1, I_2 \right) &= \frac{1}{n}\sum_{x,y} \left| I_1(x,y) - I_2(x,y) \right| \\
		d_{L2} \left( I_1, I_2 \right) &= \frac{1}{n}\sqrt{\sum_{x,y} \left( I_1(x,y) - I_2(x,y) \right)^2}  \\
		d_{ss} \left( I_1, I_2 \right) &= \frac{1}{2} \left( 1 - \SSIM \left( I_1, I_2 \right) \right) \textnormal{,}
	\end{split}
\end{align}
where $n$ is the number of pixels in the images. 

The total loss, the Gradient-Appearance Consistency loss $ \loss_{\GAC}$, is given by
\begin{gather} \label{equ:discoman_mac}
	\loss_{\GAC} = \loss_{AS} + \loss_{SA} + \loss_{A} + \loss_{S}.
\end{gather}
This loss ensures the transformation consistency of images between the two domains $ A $ and $ S $ by the generators.

Now, for making our framework adversarial, we consider two discriminators $ D_{A} $ and $ D_{S} $ that distinguish between the transformed images and the real images from domains $ A $ and $ S $ respectively. We use the following adversarial losses:
\begin{align}
	\begin{split}
		\loss_{D_{S}} &= \resizebox{.7\hsize}{!}{$ \EX\nolimits_{s^{t-T} \sim p_{s}} \left[ \log D_{S} \left( s^{t-T} \right) \right] + \EX\nolimits_{a^{t} \sim p_{a}} \left[ \log \left( 1 - D_{S} \left( G_{S} \left( a^{t} \right) \right) \right) \right] $} \\
		\loss_{D_{A}} &= \resizebox{.7\hsize}{!}{$ \EX\nolimits_{a^{t} \sim p_{a}} \left[ \log D_{A} \left( a^{t} \right) \right] + \EX\nolimits_{s^{t-T} \sim p_{s}} \left[ \log \left( 1 - D_{A} \left( G_{A} \left( s^{t-T} \right) \right) \right) \right] $} \textnormal{,}
	\end{split}
\end{align}
in which $ p_{a} $ and $ p_{s} $ describe the distributions of the input images $ a^{t} $ and $ s^{t-T} $ respectively. In theory, $ G_{S} $ and $ G_{A} $ try to generate realistic spatial gradient and appearance images by minimizing $ \loss_{D_{S}} $ and $ \loss_{D_{A}} $ accordingly. Conversely, $ D_{S} $ and $ D_{A} $ try to discriminate the real images against the generated ones by maximizing $ \loss_{D_{S}} $ and $ \loss_{D_{A}} $.

Consequently, by incorporating the reconstruction and adversarial losses, we obtain the full objective of GARDiN as follows:
\begin{gather}
	\loss_{\GARDiN} = \loss_{\GAC} + \loss_{D_{S}} + \loss_{D_{A}} \textnormal{.}
\end{gather}
This enables the model to predict the past spatial gradient of an object by looking at the current appearance and predict the current appearance by looking at the past gradient. Therefore, we hypothesize that, when an anomaly occurs, one or both of the generators will make an incorrect prediction, thus allowing its detection.

\subsubsection{Architecture}
The network architecture of our GAN is based on DiscoGAN~\cite{Kim2017}. Both generators ($ G_{S} $ and $ G_{A} $) share the same U-net like architecture using skip-connections. Considering the input image size ($ 64 \times 64 \times 1 $) for each generator of GARDiN, the encoder part is composed of 6 2-strided convolutional layers followed by the decoder  made out of 6 2-strided transpose convolutional layers, outputting an image of size ($ 64 \times 64 \times 1 $) using Sigmoid activation. The numbers of filters in the encoder are $ \left\{ 32, 64, 128, 256, 256, 256 \right\} $ and in the decoder $ \left\{ 256, 256, 128, 64, 32, 1 \right\} $. Similarly, both discriminators ($ D_{S} $ and $ D_{A} $) follow the PatchGAN architecture \cite{Isola2017} with 4 2-strided convolutional layers using $ \left\{ 32, 64, 128, 256 \right\} $ filters and a 1-strided convolutional output layer, which produces an output of size ($ 4 \times 4 \times 1 $) allowing to classify by overlapping patches. During the elaboration of our method, we noticed that using $ 4 \times 4 $ filters for all convolutional layers gives the best results. To tackle the issue of vanishing gradients, we incorporate leaky ReLU after each convolutional layer, except the last one, followed by instance normalization~\cite{Ulyanov2016} between each convolutional layer.

\subsection{Partial Mean Squared Reconstruction Errors (PMSRE)}
Once we have trained our generators $ G_{S} $ and $ G_{A} $ for predicting object-centric spatial gradient and appearance images, the next step is to compute the reconstruction errors that will be used for classification in order to predict anomalies. As previously illustrated in figure~\ref{fig:our_method}, we noticed that, most of the time, an anomaly occurs locally in one or more spatial locations in the reconstructed gradient and/or appearance images. Thus, we found it more appropriate to perform the classification on partial reconstruction errors instead of on a global one. Empirically, in both the gradient and appearance domains, we observed that dividing the pixel-level reconstruction errors into 4 blocks rendered the best performance when training our adversarial binary classifier. With an input appearance image $ a $ of an object and the predicted appearance image $ a^{*} $, we get the following partial mean squared error for an image block $B_k$:
\begin{align}
	\begin{split}
		e_{k} (a, a^{*}) &= \frac{1}{h \cdot w} \sum_{i=1}^h \sum_{j=1}^w \left( a_{ij} - a^{*}_{ij} \right)^{2} \\
	\end{split}
\end{align}
where $ h = w = 32 $ is the size of a block. By following the same logic for the input spatial gradient image $ s $ and the predicted one $ s^{*} $, we obtain the one dimensional partial reconstruction errors vector $ e $:
\begin{gather} \label{equ:e_PRE}
	\resizebox{0.9\textwidth}{!}{%
	$ e = \left[ e_{1} (a, a^{*}), e_{1} (s, s^{*}), e_{2} (a, a^{*}), e_{2} (s, s^{*}), e_{3} (a, a^{*}), e_{3} (s, s^{*}), e_{4} (a, a^{*}), e_{4} (s, s^{*}) \right] \textnormal{.} $
	}
\end{gather}

\subsection{Adversarial classification of the PMSRE}
Inspired by ALREC~\cite{Roy2019a}, we incorporate the idea of adversarially training a binary discriminator that learns to discriminate real $ e $ against fake ones generated by a generator which, in turn, learns to generate realistic $e$. In the inference mode, only the discriminator is used for predicting whether $ e $ of an object is normal or not.

\subsubsection{Formulation}
The main idea behind the use of a GAN for detecting abnormal $ e $ is to learn the distribution of normal $e$.  The generator $ G $, using an input Gaussian noise $ z $, should be able to generate realistic normal $e$ at the end of the training. In the learning process, $e$ will become more and more realistic, and we assume that this should allow the discriminator $D$ to learn the boundary between normal and abnormal $e$.  Therefore, the discriminator $ D $ should learn to classify the close to be real generated samples as fake while ensuring the detection of the real ones as real. We use the Focal Loss, $ \FL$ , for imposing more weight on harder samples than on the easier ones \cite{Lin2017a}:
\begin{gather}
	\FL (p) = - \alpha \left( 1 - p \right)^{\gamma} \log (p) \textnormal{,}
\end{gather}
where $ p $ is the prediction probability depending on the ground-truth label, $ \alpha $ affects the offset for class imbalance and $ \gamma $ helps to adjust the level of focus on hard samples. Using this, the full objective function for training our adversarial binary classifier using normal $ e $ examples is the following:
\begin{gather}
	\loss_{C} = \EX\nolimits_{e \sim p_{e}} \left[ \FL \left( D \left( e \right) \right) \right] + \EX\nolimits_{z \sim p_{z}} \left[ \FL \left( 1 - D \left( G \left( z \right) \right) \right) \right] \textnormal{.}
\end{gather}

We use label 0 for identifying fake/generated $e$ and 1 for the real/normal $ e $. In inference mode, we directly apply the trained $ D $ to produce region-level abnormality score $ s_{e} = 1 - D(e) $, which is a prediction probability that varies between $ 0 $ (normal) and $ 1 $ (abnormal). Following the same experimental protocol as \cite{Liu2018}, we normalize $ s_{e} $ scores between $ 0 $ and $ 1 $ for each sequence independently.

\subsubsection{Architecture}
We are using an architecture based on a fully-connected neural network, which enables the learning of complex pattern of $e$. The generator $ G $, taking an input noise of size 16, is composed of 5 dense hidden layers with $ \left\{ 64, 128, 128, 256, 256 \right\} $ units respectively. The discriminator $ D $ is made out of 5 dense hidden layers with respective $ \left\{ 256, 256, 128, 128, 64 \right\} $ units. In addition, to avoid overfitting, leaky ReLU and Dropout are used between each hidden layer, and the output layer follows a Sigmoid activation.

\subsection{Abnormal Events Detection}
The last step of our proposed method is to convert the region-level anomaly detection to the frame-level, to find the subsequences of a video, if any, that contain an anomaly. To obtain the frame-level anomaly score $ s_{f} $, we simply take the region-level anomaly score $ s_{e} $ which produces the maximum value. As in \cite{Ionescu2019}, we also use a Gaussian filtering technique with a standard deviation of $ 10 $ for smoothing the frame-level scores temporally throughout the sequence. A threshold can then be used to determine whether the frames are normal or not.

\section{Experiments}

\subsection{Datasets and evaluation procedure}
We conducted experiments on four publicly available datasets with varying definition and complexity of anomalies: UMN~\cite{Mehran2009}, UCSD Pedestrian~\cite{Mahadevan2010}, CUHK Avenue~\cite{Lu2013} and ShanghaiTech~\cite{Luo2017}. For all datasets, the training videos are assumed to be normal. UMN features 11 videos with 3 different scenes where anormal events are people running. We used the normal portion of 6 videos, 2 videos per scene, as training set and all the videos for the testing set as done by previous works. UCSD Pedestrin comprises two datasets: Ped1 and Ped2. Ped1 is composed of 34 training and 36 testing videos, with 40 abnormal events. Ped2 is made of 16 training and 12 testing videos, with 12 anomalies. Anomalies are the presence of skateboarders, cyclist, wheelchairs and vehicles in the pedestrian walkway areas. CUHK Avenue is composed of 16 training and 21 testing videos, where the test set contains 47 anomalies involving person running, loitering and leaving/throwing objects. Finally, ShanghaiTech is a highly challenging anomaly dataset with 13 different scenes involving diverse viewpoints and illuminations, resulting in a total of 330 training and 107 testing videos. Globally, there are 130 abnormal events in the test set with numerous types of anomalies, like people fighting, a person jumping, robbing, cyclists, etc. 

To evaluate our method, we adopted the frame-level Area Under Curve (AUC) metric. To do so, we apply the Receiver Operation Characteristic (ROC) on the frame-level anomaly ground-truth labels with respect to our predicted frame-level anomaly scores $ s_{f} $ by progressively modifying the classification threshold. As in \cite{Liu2018}, to compute the global dataset AUC score, we first got the frame-level scores per video by applying our trained method, then we combined all the scores temporally and, lastly, we computed the AUC on the concatenated scores.

\subsection{Experimental Setup}
Our method is implemented using Python 3 and Keras. For detecting multiple objects in the frames using CenterNet~\cite{zhou2019objects}, we used the model provided for the Hourglass-104 backbone with the pretrained weights from the MS-COCO dataset~\cite{Lin2014}, providing 81 different object classes. To reduce missing detections, we allow all classes with a confidence of at least $ 0.3 $.

We train GARDiN for $ 200 $ epochs of randomly shuffled mini-batches of size $ 64 $ using a learning rate starting from $ 10^{-2} $ and following a polynomial decay of power $ 2 $ every $ 25 $ epochs. To train the anomaly classifier, we follow the same training mechanism, but with a starting learning rate of $ 10^{-4} $ decaying every $ 10 $ epochs for a maximum of $ 50 $ epochs of randomly shuffled mini-batches of $ 256 $ samples. For the classifier's Focal Loss, we have empirically chosen $ \alpha = 0.1 $ and $ \gamma = 10 $. To train both frameworks, we use Adam optimizer with $ \beta_{1} = 0.5 $ and $ \beta_{2} = 0.999 $. To stabilize the adversarial training process for both GANs, we slightly smoothed the labels when training the discriminators with real samples.

\begin{table}[t]
	\centering
	\caption{Frame-level abnormal event detection AUC results (in \%) on four dataset. * Means that results were recalculated to follow the procedure in \cite{Liu2018}.}
	\label{tab:results}
	\begin{tabular}{p{3.5cm}|P{1.6cm}P{1.6cm}P{1.6cm}P{1.6cm}P{1.6cm}}
		\hline
		Method                  & UMN & Ped1 & Ped2 & Avenue & ST \\ \hline
		Conv-AE                 \cite{Hasan2016}     &     -    & $ 81.0 $ & $ 90.0 $ & $ 70.2 $ & $ 60.9 $ \\
		Discriminative          \cite{DelGiorno2016} & $ 91.0 $ &     -    &     -    & $ 78.3 $ &     -    \\
		ConvLSTM-AE             \cite{Luo2017a}      &     -    & $ 75.5 $ & $ 88.1 $ & $ 77.0 $ &     -    \\
		Deep-Cascade            \cite{Sabokrou2017}  & $ 99.6 $ &     -    &     -    &     -    &     -    \\
		STAE-optflow            \cite{Zhao2017}      &     -    & $ \mathbf{87.1} $ & $ 88.6 $ & $ 80.9 $ &     -    \\
		Deep Conv-AEs           \cite{Ribeiro2018}   &     -    & $ 56.9 $ & $ 84.7 $ & $ 77.2 $ &     -    \\
		Future frame pred       \cite{Liu2018}       &     -    & $ 83.1 $ & $ 95.4 $ & $ 84.9 $ & $ 72.8 $ \\
		OC Conv-AEs*       \cite{Ionescu2019,OC_VAD} & $ 99.6 $ &     -    &     -    & $ 86.6 $ & $ 78.6 $ \\
		M-A Correspond          \cite{Nguyen2019a}   &     -    &     -    & $ 96.2 $ & $ 86.9 $ &     -    \\ \hline
		GARDiN-VAD (ours)                          & $ \mathbf{99.7} $ & $ 85.2 $ & $ \mathbf{97.5} $ & $ \mathbf{87.3} $ & $ \mathbf{81.1} $ \\ \hline
	\end{tabular}
\end{table}
\begin{figure*}[!t]
	\centering
	\subfloat[Ped1\label{fig:results_ped1}]{
		\includegraphics[width=0.45\textwidth]{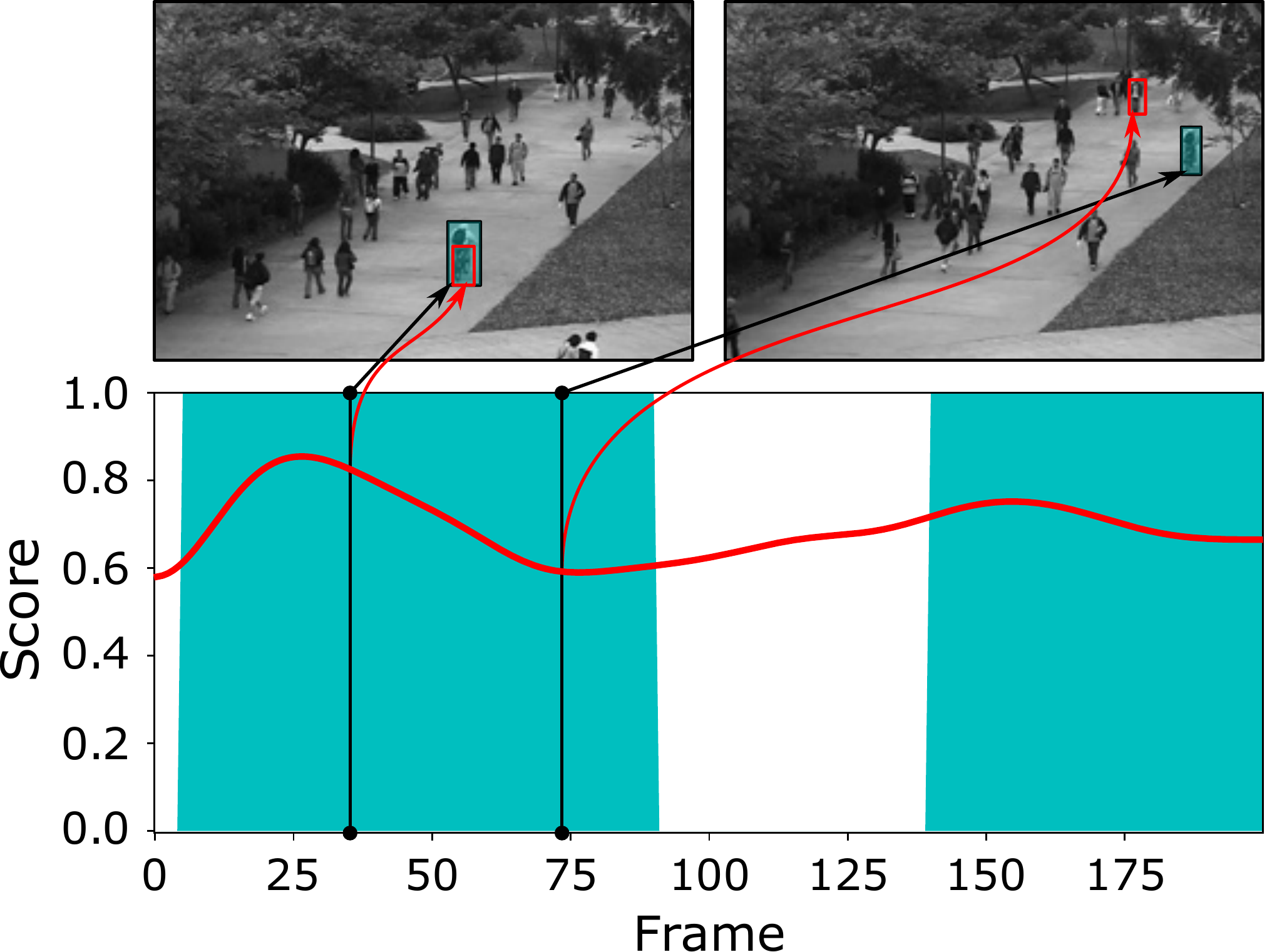}
	} 
	\subfloat[Ped2\label{fig:results_ped2}]{
		\includegraphics[width=0.45\textwidth]{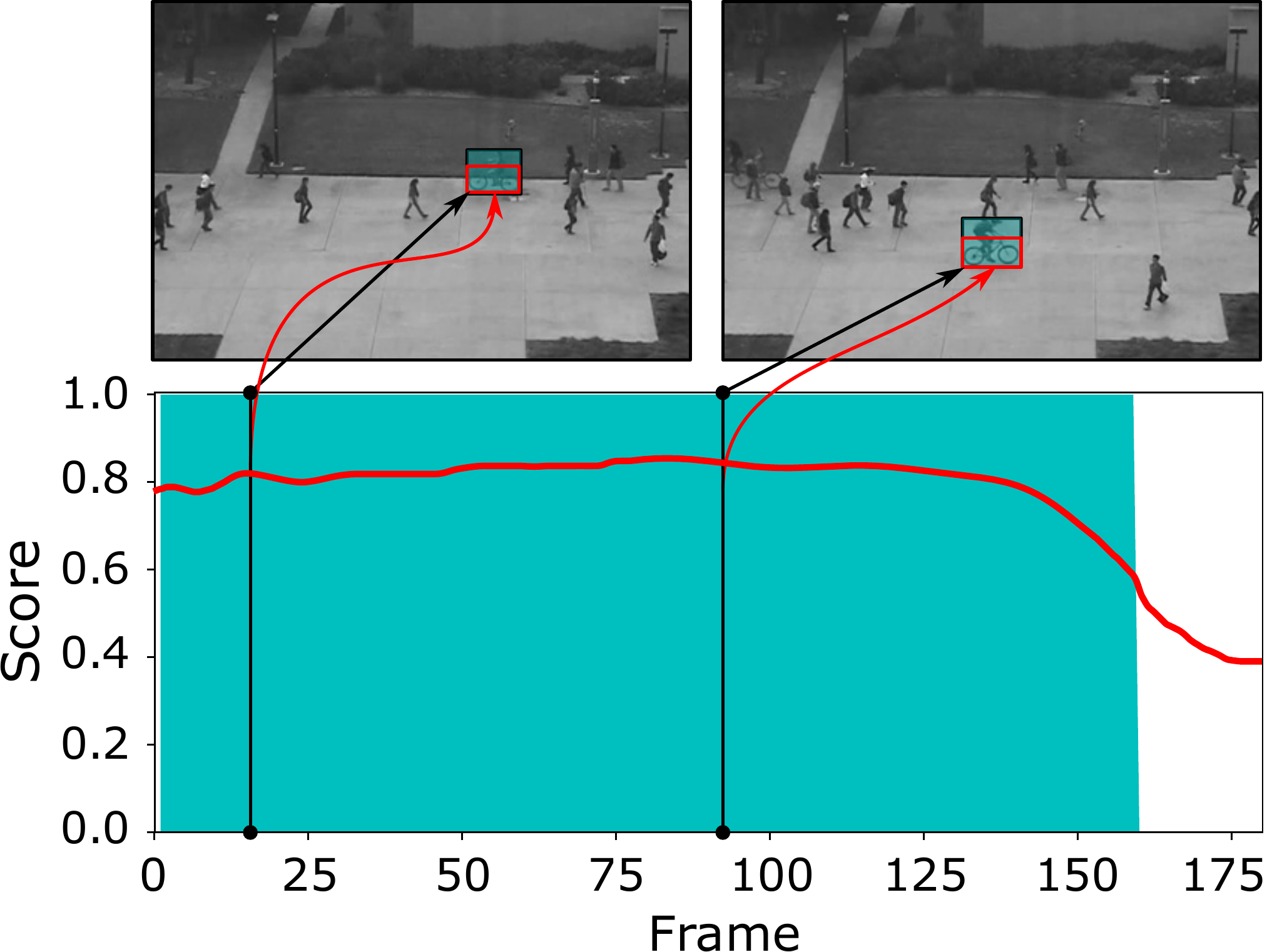}
	}\\
	\subfloat[Avenue\label{fig:results_avenue}]{
		\includegraphics[width=0.45\textwidth]{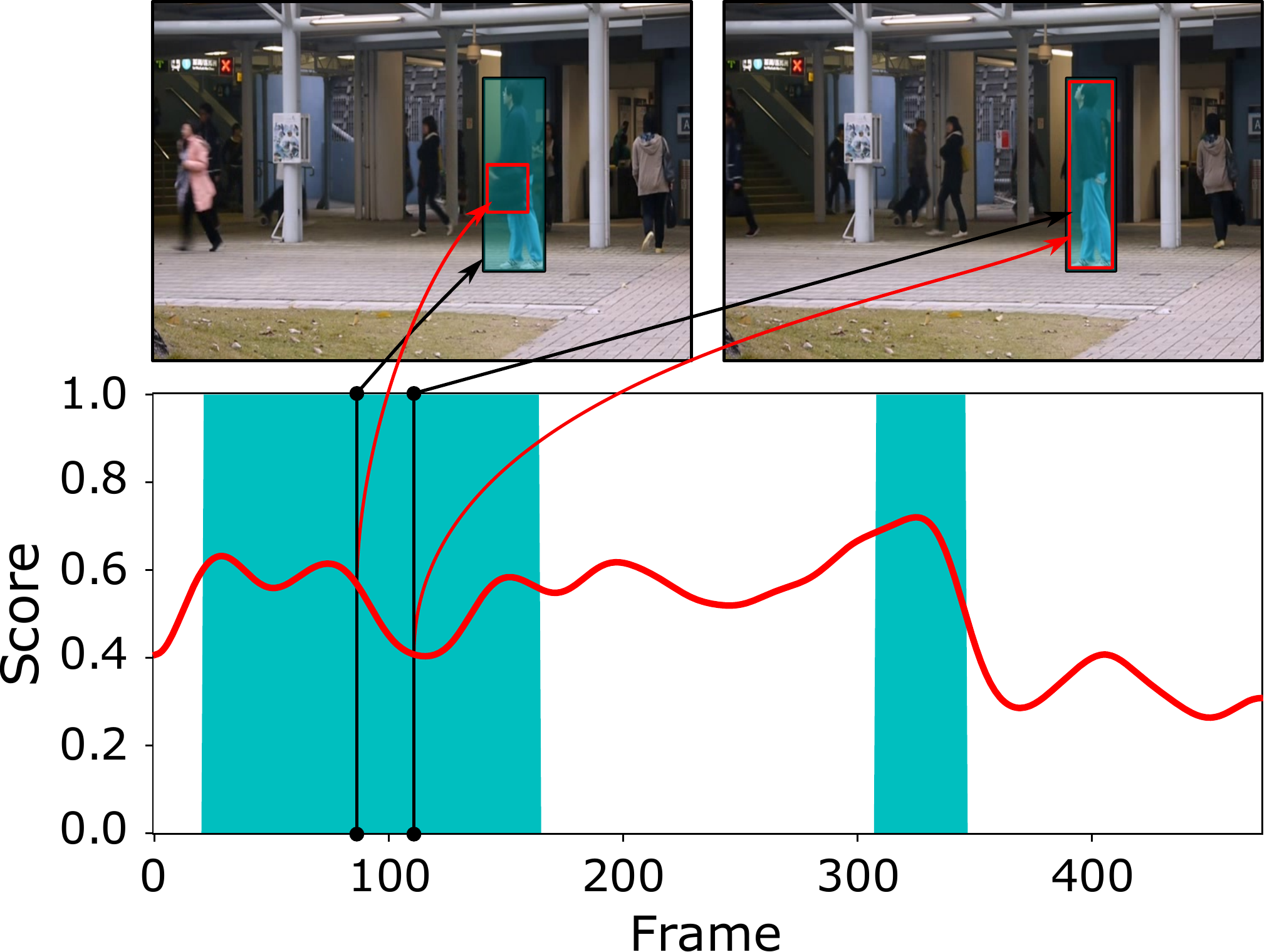}
	}
	\subfloat[ShanghaiTech\label{fig:results_shanghai}]{
		\includegraphics[width=0.45\textwidth]{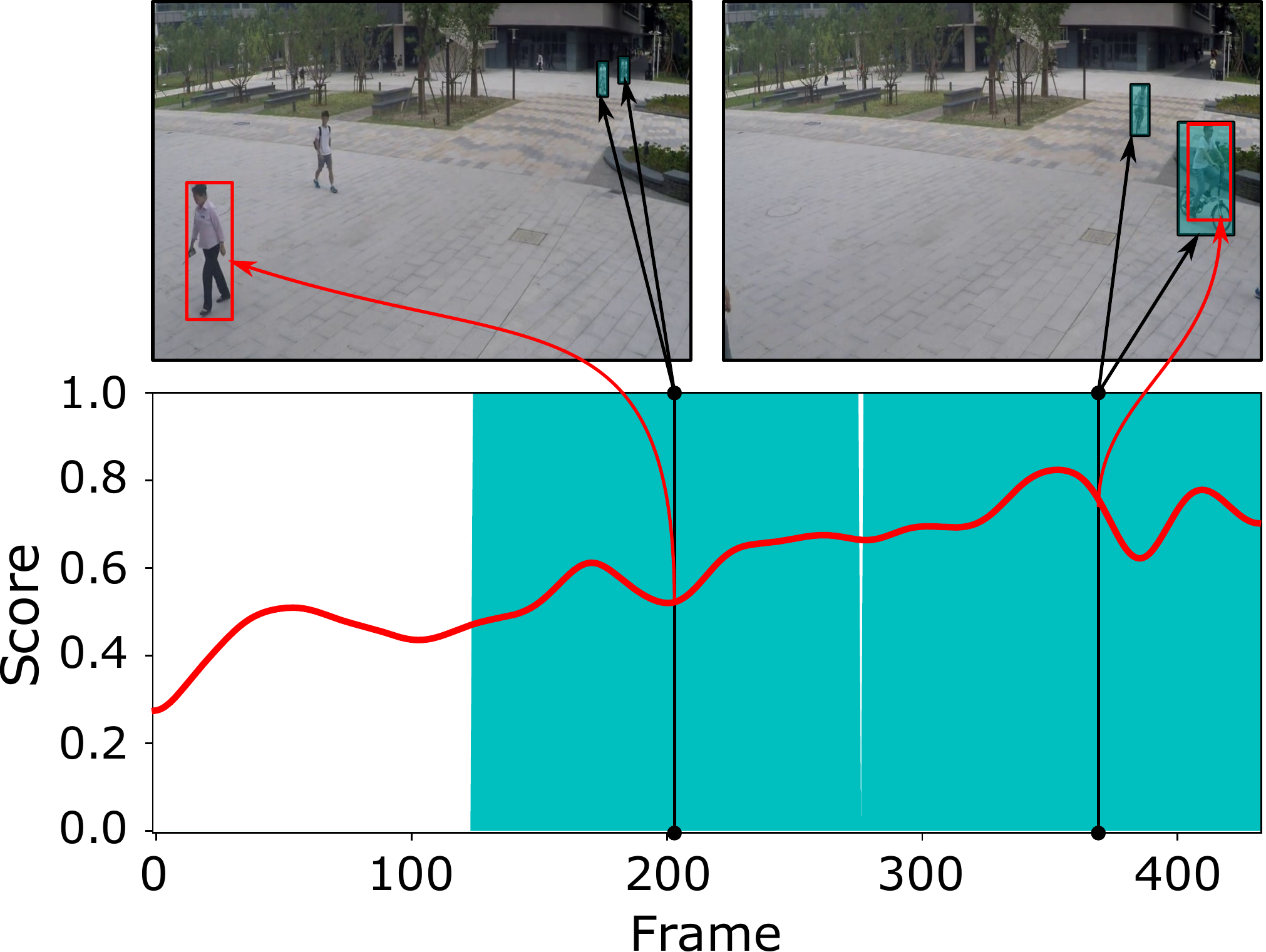}
	}
	\caption{Qualitative anomaly detection results on test videos from each dataset. Red curves show the frame-level anomaly score and the areas in cyan represent the ground-truth abnormal frames. Black and red arrows point to the ground-truth bounding boxes and the detected regions by GARDiN-VAD respectively.}
	\label{fig:results}
\end{figure*}
\subsection{Results}
Table~\ref{tab:results} presents our frame-level AUC anomaly detection results on the four datasets. We also included in the table some recent state-of-the-art methods evaluated on at least one of the considered datasets. Figure~\ref{fig:results} illustrates anomaly detections on Ped1, Ped2, Avenue and ShanghaiTech using our proposed method.

\subsubsection{UMN}
For the UMN dataset, our approach significantly outperforms \cite{DelGiorno2016} and is on par with \cite{Sabokrou2017,Ionescu2019} by achieving a near perfect result. More specifically, our proposed GARDiN-VAD can accurately detect the people escape instances on all three different scenarios, while discarding the background information. This illustrates the applicability of our object-centric adversarial approach for detecting real-world crowd panic events.

\subsubsection{UCSD Pedestrian}
On Ped1, we note a notable improvement in the AUC score using GARDiN-VAD compared to some anomaly detection counterparts \cite{Luo2017a,Ribeiro2018} and almost on par results with \cite{Mahadevan2010,Hasan2016,Liu2018}. However, we notice that the spatio-temporal-based method of \cite{Zhao2017} largely surpasses GARDiN-VAD. In fact, the spatio-temporal auto-encoder in \cite{Zhao2017} was only evaluated on videos having a single scene and, unlike our method, they are not applicable to datasets which contains various scenes. Even though the anomaly definition on Ped1 is simple, the fact that the image resolution is only $ 158 \times 238 $ makes the input data for GARDiN significantly more noisy compared to other datasets. Moreover, as illustrated in figure~\ref{fig:results}\subref{fig:results_ped1}, there might be heavy occlusions in some portions of the crowded scene, thus making it difficult for our detector to extract a well defined region, especially for the far top-right objects. Nevertheless, on Ped2, we obtain the best performance, outperforming all  state-of-the-art methods, even the recent ones \cite{Liu2018,Nguyen2019a}. In fact, as shown in figure \ref{fig:results}\subref{fig:results_ped2}, there are noticeably less occlusions than Ped1 and the foreground objects are visually clearer. Despite having a better resolution, Ped2 is easier than Ped1 because of the lateral viewpoint.

\subsubsection{CUHK Avenue}
For the Avenue dataset, our method competes with \cite{Liu2018,Ionescu2019,Nguyen2019a} and significantly outperforms others, which shows that object-centric-based methods can be more robust to occlusions and camera jittering. However, it cannot detect anomalies involving the interaction between multiple objects in the scene, as illustrated by a person throwing a bag which goes outside the video frame in figure \ref{fig:results}\subref{fig:results_avenue}, which is expected because this is not a local anomaly. Nevertheless, the frame-level AUC results show that object-centric approaches perform overall well for detecting local anomalies on a challenging side-view scenario.

\subsubsection{ShanghaiTech}
Lastly, on the most challenging dataset ShanghaiTech, our method noticeably outperforms the method of \cite{Hasan2016} by an absolute gain of around 20\% and it obtains results better than \cite{Liu2018}. It also performs slightly better than the other object-centric method~\cite{Ionescu2019,OC_VAD} while relying only on the past and present observations. Although sometimes, depending on the camera angle, cyclists can have lower anomaly scores than pedestrians as shown in figure~\ref{fig:results}\subref{fig:results_shanghai}, methods relying only on the regions of the objects will be able to detect local anomalies across different scenes, mainly due to the fact that they exclude background information and are less context-dependent.
\begin{table}[t]
	\centering
	\caption{Ablation study AUC results (in \%) on Ped2.}
	\label{tab:ablation_study}
	\subfloat[GARDiN losses using One-Class SVM with L2 distance metric.]{
		\label{tab:loss}
		\makebox[0.9\linewidth]{\begin{tabular}{P{1.3cm}|P{1.82cm}P{1.82cm}P{1.82cm}P{1.82cm}}
				\hline
				$ \loss_{AM} $ & \checkmark & \checkmark & \checkmark & \checkmark \\
				$ \loss_{MA} $ &     ---    & \checkmark & \checkmark & \checkmark \\
				$ \loss_{A} $  &     ---    &     ---    & \checkmark & \checkmark \\
				$ \loss_{M} $  &     ---    &     ---    &     ---    & \checkmark \\ \hline
				AUC            &  $ 79.3 $  &  $ 81.2 $  &  $ 82.7 $  &  $ \mathbf{85.2} $  \\ \hline
		\end{tabular}}
	} \\
	\subfloat[Distance metrics in GARDiN losses using One-Class SVM.]{
		\label{tab:distance}
		\makebox[0.9\linewidth]{\begin{tabular}{P{1.3cm}|P{1cm}P{1cm}P{1cm}P{1cm}P{1cm}P{1cm}P{1cm}}
				\hline
				$ d_{L_{1}} $ & \checkmark & \checkmark &     ---     &     ---     & \checkmark & \checkmark & \checkmark \\
				$ d_{L_{2}} $ &     ---    & \checkmark &     ---     &     ---     & \checkmark & \checkmark & \checkmark \\
				$ d_{ss} $    &     ---    &     ---    & \checkmark  &     ---     & \checkmark &     ---    & \checkmark \\
				$ d_{nr} $  &     ---    &     ---    &     ---     & \checkmark  &     ---    & \checkmark & \checkmark \\ \hline
				AUC           &  $ 85.2 $  &  $ 87.1 $  &  $ 84.5 $   &  $ 83.6 $   &  $ \mathbf{91.3} $  &  $ 88.7 $  &  $ 89.8 $  \\ \hline
		\end{tabular}}
	} \\
	\subfloat[Classification methods of reconstruction errors.]{
		\label{tab:classification}
		\makebox[0.9\linewidth]{\begin{tabular}{p{3.4cm}|P{1.3cm}P{1.3cm}P{1.3cm}P{1.3cm}}
				\hline
				Method								&     L1     &     L2     &    SSIM    &   PMSRE   \\ \hline
				OC-SVM \cite{Scholkopf2001}         &  $ 88.9 $  &  $ 91.3 $  &  $ 92.1 $  &  $ 93.4 $  \\
				One-vs-rest-SVMs \cite{Ionescu2019} &  $ 89.4 $  &  $ 92.1 $  &  $ 92.8 $  &  $ 94.3 $  \\
				DAE \cite{Roy2018a}                 &     ---    &     ---    &     ---    &  $ 95.2 $  \\
				ALREC \cite{Roy2019a}               &     ---    &     ---    &     ---    &  $ 95.9 $  \\
				ALREC-FL (ours)                     &     ---    &     ---    &     ---    &  $ \mathbf{97.5} $  \\ \hline
		\end{tabular}}
	}
\end{table}

\begin{table}[t]
	\centering
	\caption{GARDiN AUC results (in \%) on UCSD (Ped1 and Ped2), Avenue and ShanghaiTech using one of two different detectors.}
	\label{tab:detector_study}
	\begin{tabular}{p{2.3cm}|P{2cm}P{2cm}P{2cm}P{2cm}}
		\hline
		Detector        & Ped1 & Ped2 & Avenue & ShanghaiTech \\ \hline
		RetinaNet \cite{Lin2017a}        & $ 83.6 $ & $ 97.4 $ & $ 83.1 $ & $ 80.3 $ \\
		CenterNet \cite{zhou2019objects} & $ \mathbf{85.2} $ & $ \mathbf{97.5} $ & $ \mathbf{87.3} $ & $ \mathbf{81.1} $ \\ \hline
	\end{tabular}
\end{table}
\subsection{Ablation Study}
We chose Ped2 for conducting our ablation study, presented in table~\ref{tab:ablation_study}, since we can train our models faster on it and the anomaly definition generalizes well across other datasets. 
First, as summarized in table~\ref{tab:ablation_study}\subref{tab:loss}, to validate the loss function $ \loss_{MAC} $ in equation (\ref{equ:discoman_mac}) which ensures object-centric gradient and appearance consistency when training GARDiN, we used the L2 distance measure between real and generated images in the loss functions $ \loss_{AM} $, $ \loss_{MA} $, $ \loss_{A} $ and $ \loss_{M} $, and used a simple outlier detection technique based on a One-Class SVM for detecting anomalies. By testing various combinations, we observe that the fusion of all the losses produce the best result. Secondly, to find the most appropriate distance metric that will be used in $ \loss_{MAC} $, we also tested several combinations of distances based on L1 ($ d_{L_{1}} $), L2 ($ d_{L_{2}} $), SSIM ($ d_{ss} $) and PSNR inspired by \cite{Liu2018}($ d_{nr} $). Table \ref{tab:ablation_study}\subref{tab:distance} shows the importance of combining image quality distance metrics for better assessment of the correspondence between gradient and appearance of objects. Lastly, we tested several unsupervised classification approaches \cite{Scholkopf2001,Ionescu2019,Roy2018a,Roy2019a} for distinguishing normal and abnormal reconstruction errors of generated images from GARDiN. The AUC results in table \ref{tab:ablation_study}\subref{tab:classification} confirm that the adversarial method outperforms others by a large margin. We observe that incorporating Focal Loss \cite{Lin2017a} when training ALREC \cite{Roy2019a} (ALREC-FL) notably improves the performance for abnormal PMSRE detection.

We also evaluated the impact of the object detector on the overall AUC results. We compared CenterNet \cite{zhou2019objects} and RetinaNet \cite{Lin2017a} for this task. Table \ref{tab:detector_study} shows that using a higher performance detector (CenterNet) noticeably improves the anomaly detection performance, mostly on Ped1 and Avenue datasets which contain the highest amount of occlusions and noise among the studied datasets.

\subsection{Inference Running Time}
On a Intel i5-9400F machine with 16 GB RAM using Nvidia RTX 2070 GPU with 8 GB VRAM and considering an average number of objects in a video frame of $ 5 $, the preprocessing step for the detection of objects and the extraction of their gradient images takes about $ 75 $ ms per frame. In inference mode, the running time of the combined GARDiN and ALREC-FL frameworks is approximately $ 5 $ ms per frame. Thus, the overall pipeline of our proposed method consumes roughly $ 80 $ ms for a single frame, leading to a running speed of $ 12.5 $ FPS.

\section{Conclusion}
In this paper, we propose GARDiN-VAD: a novel unsupervised approach for local anomaly detection in videos based on object-centric adversarial learning trained using normal training samples only. First, we extract the appearance and the gradient of all the objects in the scenes by using the pretrained CenterNet object detector. Then, we train GARDiN, composed of two generators and two discriminators to learn the relationship between appearance and gradient. After that, we train ALREC-FL with PMSRE to classify abnormal PMSRE caused by abnormal appearance-gradient relationships. On four public benchmarks, our method yields competitive  results, superior to state-of-the-art approaches.
%
%
%
\bibliographystyle{splncs04}
\bibliography{bib/paper}
\end{document}